\documentclass{article}
\usepackage[accepted]{icml2016}
\usepackage{times}
\usepackage{graphicx} 
\usepackage{epsfig} 
\usepackage{subfigure} 
\usepackage{array}
\usepackage{color}
\newcolumntype{L}{>{\small\centering\arraybackslash}m{1.5cm}}
\usepackage{natbib}
\usepackage{amsbsy}
\usepackage{hyperref}

\icmltitlerunning{LSTM-based Encoder-Decoder for Multi-sensor Anomaly Detection}

\begin{document} 

\twocolumn[
\icmltitle{LSTM-based Encoder-Decoder for Multi-sensor Anomaly Detection  }

\icmlauthor{Pankaj Malhotra, Anusha Ramakrishnan, Gaurangi Anand, Lovekesh Vig, Puneet Agarwal, Gautam Shroff\\}{\{malhotra.pankaj, anusha.ramakrishnan, gaurangi.anand, lovekesh.vig, puneet.a, gautam.shroff\}@tcs.com}
\icmladdress{TCS Research, New Delhi, India}

\icmlkeywords{Anomaly Detection,LSTM Encoder Decoder}

\vskip 0.2in
]

\begin{abstract} 
Mechanical devices such as engines, vehicles, aircrafts, etc., are typically instrumented with numerous sensors to capture the behavior and health of the machine. However, there are often external factors or variables which are not captured by sensors leading to time-series which are inherently unpredictable. For instance, manual controls and/or unmonitored environmental conditions or load may lead to inherently unpredictable time-series. Detecting anomalies in such scenarios becomes challenging using standard approaches based on mathematical models that rely on stationarity, or prediction models that utilize prediction errors to detect anomalies. 
We propose a Long Short Term Memory Networks based \textbf{Enc}oder-\textbf{Dec}oder scheme for \textbf{A}nomaly \textbf{D}etection (EncDec-AD) that learns to reconstruct `normal' time-series behavior, and thereafter uses reconstruction error to detect anomalies. We experiment with three publicly available quasi predictable time-series datasets: power demand, space shuttle, and ECG, and two real-world engine datasets with both predictive and unpredictable behavior. \textit{We show that EncDec-AD is robust and can detect anomalies from predictable, unpredictable, periodic, aperiodic, and quasi-periodic time-series. Further, we show that EncDec-AD is able to detect anomalies from short time-series (length as small as 30) as well as long time-series (length as large as 500)}.
\end{abstract}

\section{Introduction}
\begin{figure}
    \centering
    \subfigure[Predictable\label{fig:predictable}]{\includegraphics[width=0.495\linewidth]{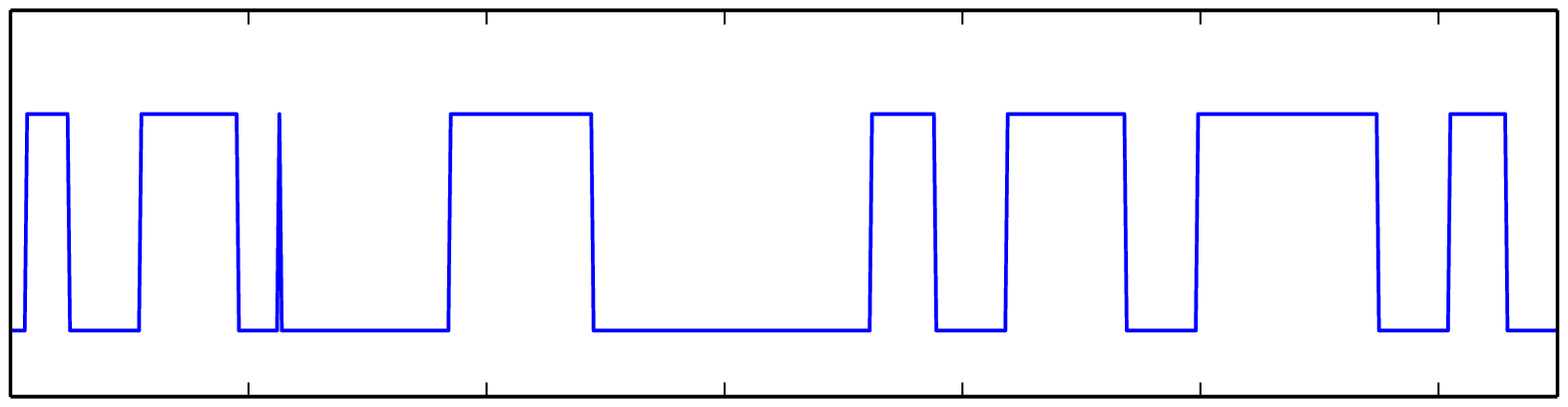}}
    \subfigure[Unpredictable\label{fig:unpredictable}]{\includegraphics[width=0.495\linewidth]{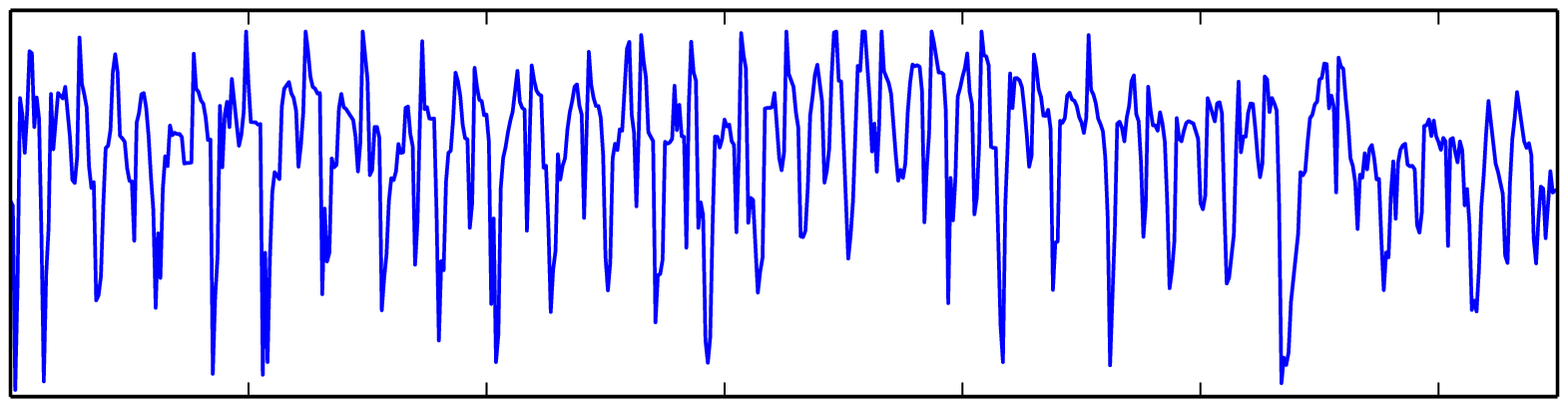}}
    \caption{\label{fig:Predictability}Readings for a manual control sensor.}
    \vspace{-14pt}
\end{figure}

In real-world sensor data from machines, there are scenarios when the behavior of a machine changes based on usage and external factors which are difficult to capture. For example, a laden machine behaves differently from an unladen machine. Further, the relevant information pertaining to whether a machine is laden or unladen may not be available. The amount of load on a machine at a time may be unknown or change very frequently/abruptly, for example, in an earth digger. A machine may have multiple manual controls some of which may not be captured in the sensor data. Under such settings, it becomes difficult to predict the time-series, even for very near future (see Figure \ref{fig:Predictability}), rendering ineffective prediction-based time-series anomaly detection models, such as ones based on exponentially weighted moving average (EWMA) \cite{stat_AD}, SVR\cite{ma2003online}, or Long Short-Term Memory (LSTM) Networks \cite{p:lstm-ad}.

LSTM networks \cite{hochreiter1997long} are recurrent models that have been used for many sequence learning tasks like handwriting recognition, speech recognition, and sentiment analysis. LSTM Encoder-Decoder models have been recently proposed for sequence-to-sequence learning tasks like machine translation \cite{p:seq2seq, p:seq2seqNIPS2014}. An LSTM-based encoder is used to map an input sequence to a vector representation of fixed dimensionality. The decoder is another LSTM network which uses this vector representation to produce the target sequence. Other variants have been proposed for natural language generation and reconstruction \cite{p:autoencoderLanguageGen}, parsing \cite{vinyals2015grammar}, image captioning \cite{p:tensorflowEncDec}.

We propose an LSTM-based \textbf{Enc}oder-\textbf{Dec}oder scheme for \textbf{A}nomaly \textbf{D}etection in multi-sensor time-series (EncDec-AD). An encoder learns a vector representation of the input time-series and the decoder uses this representation to reconstruct the time-series. The LSTM-based encoder-decoder is trained to reconstruct instances of `normal' time-series with the target time-series being the input time-series itself. Then, the reconstruction error at any future time-instance is used to compute the likelihood of anomaly at that point. We show that such an encoder-decoder model learnt using only the normal sequences can be used for detecting anomalies in multi-sensor time-series: 
The intuition here is that the encoder-decoder pair would only have seen normal instances during training and learnt to reconstruct them. When given an anomalous sequence, it may not be able to reconstruct it well, and hence would lead to higher reconstruction errors compared to the reconstruction errors for the normal sequences.

EncDec-AD uses only the normal sequences for training. This is particularly useful in scenarios when anomalous data is not available or is sparse,  making it difficult to learn a classification model over the normal and anomalous sequences. This is especially true of machines that undergo periodic maintainance and therefore get serviced before anomalies show up in the sensor readings. 
\section{EncDec-AD}
Consider a time-series $X=\{\mathbf{x}^{(1)},\mathbf{x}^{(2)}, ..., \mathbf{x}^{(L)}\}$ of length $L$, where each point $\mathbf{x}^{(i)} \in {R}^m$ is an $m$-dimensional vector of readings for $m$ variables at time-instance $t_i$. We consider the scenario where multiple such time-series are available or can be obtained by taking a window of length $L$ over a larger time-series. We first train the LSTM Encoder-Decoder model to reconstruct the normal time-series. The reconstruction errors are then used to obtain the likelihood of a point in a test time-series being anomalous s.t. for each point $\mathbf{x}^{(i)}$, an anomaly score $a^{(i)}$ of the point being anomalous is obtained. A higher anomaly score indicates a higher likelihood of the point being anomalous.

\subsection{LSTM Encoder-Decoder as reconstruction model}
We train an LSTM encoder-decoder to reconstruct instances of normal time-series. The LSTM encoder learns a fixed length vector representation of the input time-series and the LSTM decoder uses this representation to reconstruct the time-series using the current hidden state and the value predicted at the previous time-step. Given $X$, $\mathbf{h}_E^{(i)}$ is the hidden state of encoder at time $t_i$ for each $i \in \{1,2,...,L\}$, where $\mathbf{h}_E^{(i)} \in {R}^c$, $c$ is the number of LSTM units in the hidden layer of the encoder. The encoder and decoder are jointly trained to reconstruct the time-series in reverse order (similar to \cite{p:seq2seqNIPS2014}), i.e. the target time-series is $\{\mathbf{x}^{(L)},\mathbf{x}^{(L-1)}, ..., \mathbf{x}^{(1)}\}$. The final state $\mathbf{h}_E^{(L)}$ of the encoder is used as the initial state for the decoder. A linear layer on top of the LSTM decoder layer is used to predict the target. During training, the decoder uses $\mathbf{x}^{(i)}$ as input to obtain the state $\mathbf{h}_D^{(i-1)}$, and then predict $\mathbf{x'}^{(i-1)}$ corresponding to target $\mathbf{x}^{(i-1)}$. During inference, the predicted value $\mathbf{x'}^{(i)}$ is input to the decoder to obtain $\mathbf{h}_D^{(i-1)}$ and predict $\mathbf{x'}^{(i-1)}$.
The model is trained to minimize the objective $\sum_{X \in s_N}\sum_{i=1}^{L}\|\mathbf{x}^{(i)}-\mathbf{x}'^{(i)}\|^2$, where $s_N$ is set of normal training sequences.

\begin{figure}
 \centering
\includegraphics[width=\columnwidth]{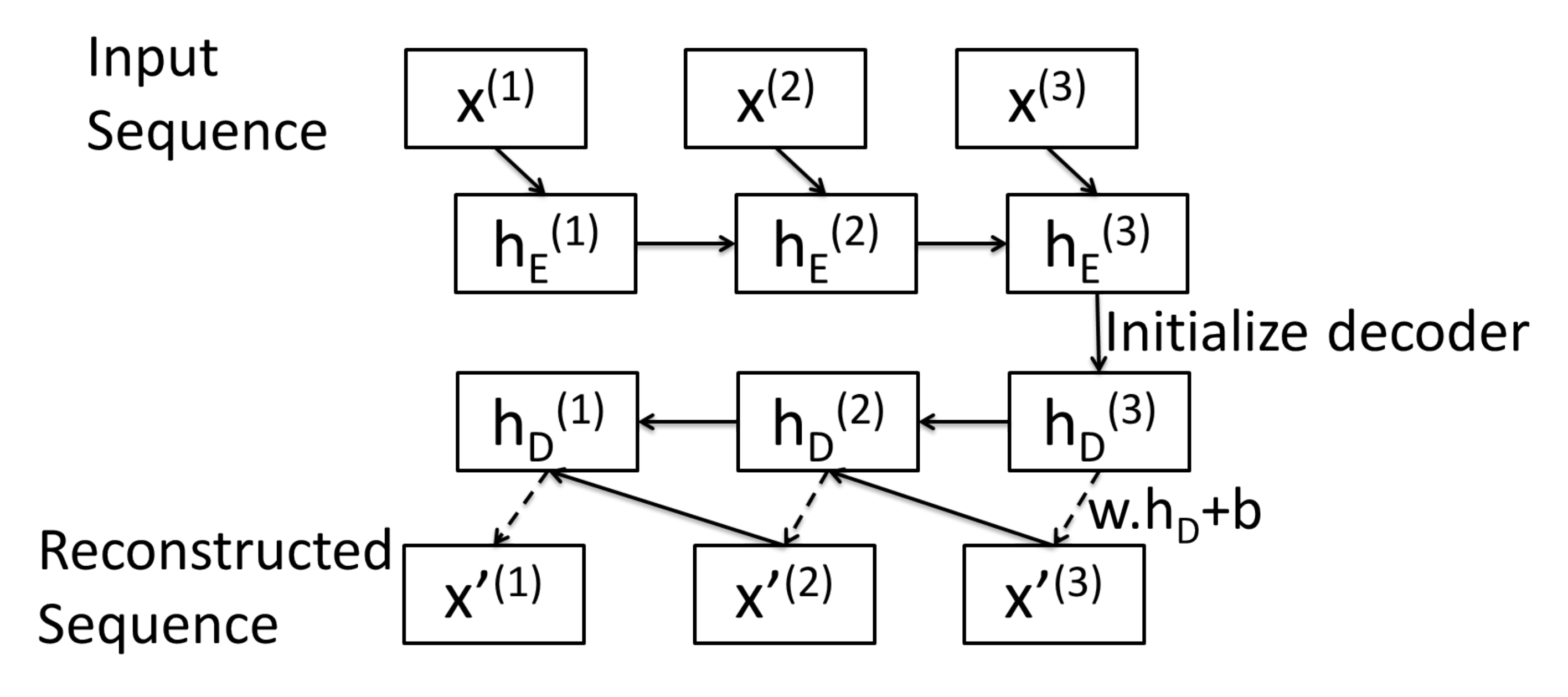}
\caption{\label{fig:EncDec}LSTM Encoder-Decoder inference steps for input $\{\mathbf{x}^{(1)},\mathbf{x}^{(2)}, \mathbf{x}^{(3)}\}$ to predict $\{\mathbf{x'}^{(1)},\mathbf{x'}^{(2)}, \mathbf{x'}^{(3)}\}$}
\vspace{-13pt}
\end{figure}

Figure \ref{fig:EncDec} depicts the inference steps in an LSTM Encoder-Decoder reconstruction model for a sequence with $L=3$. The value $\mathbf{x}^{(i)}$ at time instance $t_i$ and the hidden state $\mathbf{h}_E^{(i-1)}$ of the encoder at time $t_i-1$ are used to obtain the hidden state $\mathbf{h}_E^{(i)}$ of the encoder at time $t_i$. The hidden state $\mathbf{h}_E^{(3)}$ of the encoder at the end of the input sequence is used as the initial state $\mathbf{h}_D^{(3)}$ of the decoder s.t. $\mathbf{h}_D^{(3)}=\mathbf{h}_E^{(3)}$. A linear layer with weight matrix $\mathbf{w}$ of size $c\times m$ and bias vector $b \in R^m$ on top of the decoder is used to compute $\textbf{x}'^{(3)}=\mathbf{w^Th}_D^{(3)}+\mathbf{b}$. The decoder uses $\mathbf{h}_D^{(i)}$ and prediction $\mathbf{x'}^{(i)}$ to obtain the next hidden state $\mathbf{h}_D^{(i-1)}$.

\subsection{Computing likelihood of anomaly}\label{ssec:computeLike}
Similar to \cite{p:lstm-ad}, we divide the normal time-series into four sets of time-series: $s_N$, $v_{N1}$, $v_{N2}$, and $t_N$, and the anomalous time-series into two sets $v_A$ and $t_A$. The set of sequences $s_N$ is used to learn the LSTM encoder-decoder reconstruction model. The set $v_{N1}$ is used for early stopping while training the encoder-decoder model.
The reconstruction error vector for $t_i$ is given by $\mathbf{e}^{(i)}=|\mathbf{x}^{(i)}-\mathbf{x'}^{(i)}|$. The error vectors for the points in the sequences in set $v_{N1}$ are used to estimate the parameters $\boldsymbol{\mu}$ and $\mathbf{\Sigma}$ of a Normal distribution $\mathcal{N}(\boldsymbol{\mu},\mathbf{\Sigma})$ using Maximum Likelihood Estimation. Then, for any point $\mathbf{x}^{(i)}$, the anomaly score $a^{(i)}=(\mathbf{e}^{(i)}-\boldsymbol{\mu})^T\mathbf{\Sigma}^{-1}(\mathbf{e}^{(i)}-\boldsymbol{\mu})$.

In a supervised setting, if $a^{(i)}> \tau$, a point in a sequence can be predicted to be ``anomalous'', otherwise ``normal''. When enough anomalous sequences are available, a threshold $\tau$ over the likelihood values is learnt to maximize $F_{\beta} = (1+\beta^2)\times P\times R/(\beta^2 P+R)$, where P is precision, R is recall, ``anomalous'' is the positive class and ``normal'' is the negative class.  If a window contains an anomalous pattern, the entire window is labeled as ``anomalous''. This is helpful in many real-world applications where the exact position of anomaly is not known. For example, for the engine dataset (refer Section \ref{sec:experiments}), the only information available is that the machine was repaired on a particular date. The last few operational runs prior to repair are assumed to be anomalous and the first few operational runs after the repair are assumed to be normal. We assume $\beta < 1$ since the fraction of actual anomalous points in a sequence labeled as anomalous may not be high, and hence lower recall is expected. The parameters $\tau$ and $c$ are chosen with maximum $F_{\beta}$ score on the validation sequences in $v_{N2}$ and $v_A$.

\section{Experiments}\label{sec:experiments}
We consider four real-world datasets: power demand, space shuttle valve, ECG, and engine (see Table \ref{t:natureData}).
The first three are taken from \cite{p:keogh-hot-sax} whereas the engine dataset is a proprietary one encountered in a real-life project. The engine dataset contains data for two different applications: Engine-P where the time-series is quasi-predictable, Engine-NP where the time-series is unpredictable, for reasons such as mentioned earlier.

In our experiments, we consider architectures where both the encoder and decoder have single hidden layer with $c$ LSTM units each. Mini-batch stochastic optimization based on Adam Optimizer \cite{p:adamOpt} is used for training the LSTM Encoder-Decoder. 
Table \ref{t:results} shows the performance of EncDec-AD on all the datasets. 

\begin{table}
\resizebox{\linewidth}{!}{
\begin{tabular}{|c|c|c|c|c|c|c|} \hline
\textbf{Datasets}&\textbf{Predictable}&\textbf{Dimensions}&\textbf{Periodicity}&$\mathbf{N}$&$\mathbf{N_{n}}$&$\mathbf{N_{a}}$\\ \hline
\textbf{Power Demand}&Yes&1&Periodic&1&45&6\\ \hline
\textbf{Space Shuttle}&Yes&1 & Periodic&3&20&8\\ \hline
\textbf{Engine-P}&Yes&12 & Aperiodic&30&240&152 \\ \hline
\textbf{Engine-NP}&No&12 & Aperiodic&6&200&456\\ \hline
\textbf{ECG} &Yes&1 & Quasi-periodic&1&215&1\\ \hline
\end{tabular}
}
\caption{Nature of datasets. $N$, $N_{n}$ and $N_a$ is no. of original sequences, normal subsequences and anomalous subsequences, respectively.\label{t:natureData}}
\vspace{-10pt}
\end{table}

\begin{table}
\resizebox{\linewidth}{!}{
\begin{tabular}{|c|c|c|c|c|c|c|c|} \hline
\textbf{Datasets}&\textbf{L}&\textbf{c}&$\boldsymbol{\beta}$&\textbf{P}&\textbf{R}&\textbf{F$_{\boldsymbol{\beta}}$-score}&\textbf{TPR/FPR}\\ \hline
\textbf{Power Demand}&84& 40& 0.1&0.92 & 0.04 &0.77&33.0\\ \hline
\textbf{Space Shuttle}&500 &50 &0.05& 0.83&0.08 & 0.81&4.9\\ \hline
\textbf{Engine-P}&30 &40 &0.05& 0.94&0.02 & 0.82& 13.8 \\ \hline
\textbf{Engine-NP}&30 &90&0.05&1.0& 0.01&0.83& $\infty$ \\ \hline
\textbf{ECG} & 208&45&0.05& 1.0&0.005& 0.65 & $\infty$\\ \hline
\end{tabular}
}
\caption{F$_\beta$-scores and positive likelihood ratios (TPR/FPR).\label{t:results}}
\vspace{-10pt}
\end{table}

\begin{figure}\label{fig:Results}
    \centering
    \subfigure[Power-N\vspace{-10pt}\label{fig:power_n}]{\includegraphics[width=0.495\linewidth]{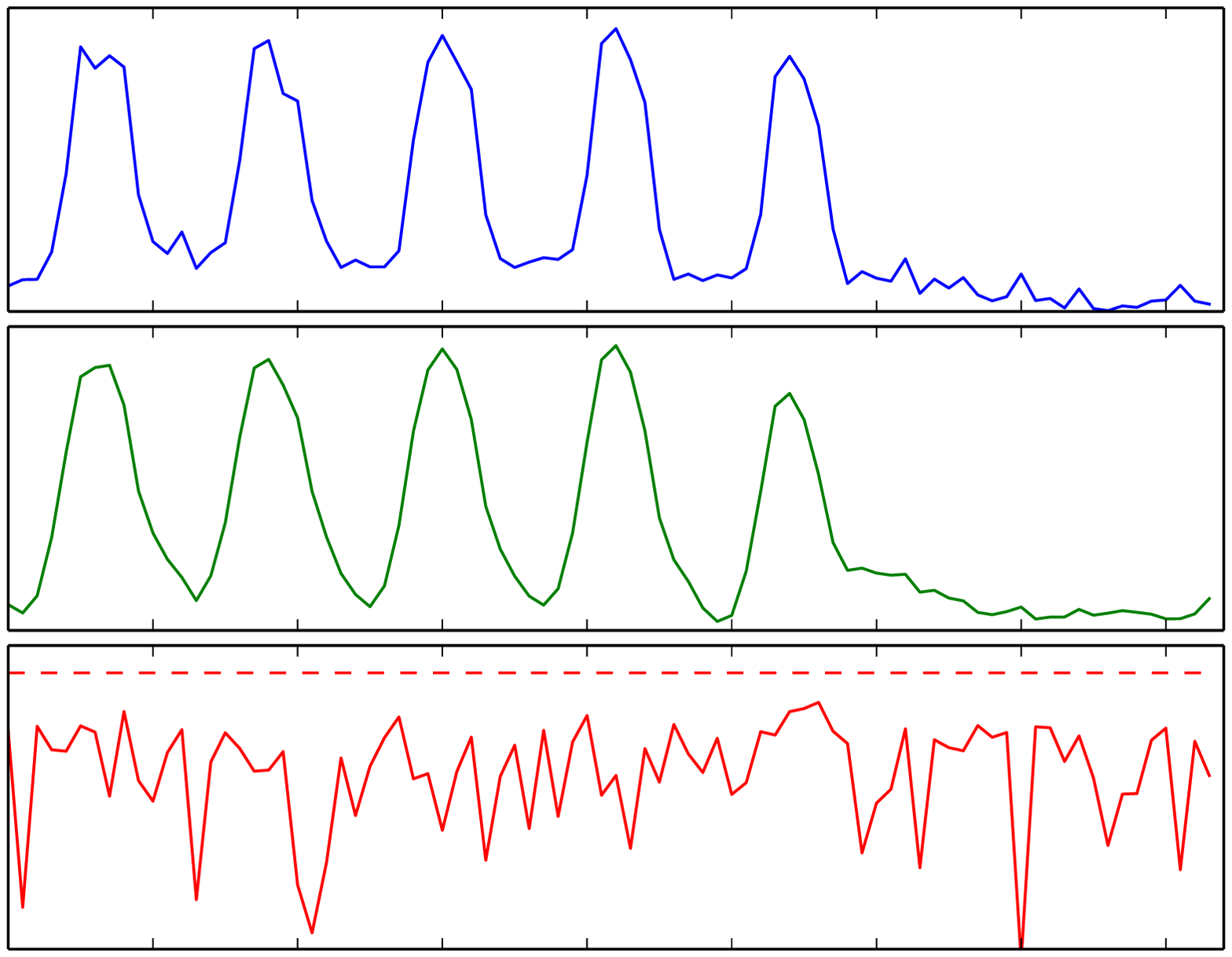}}
    \subfigure[Power-A\vspace{-10pt}\label{fig:power_a}]{\includegraphics[width=0.495\linewidth]{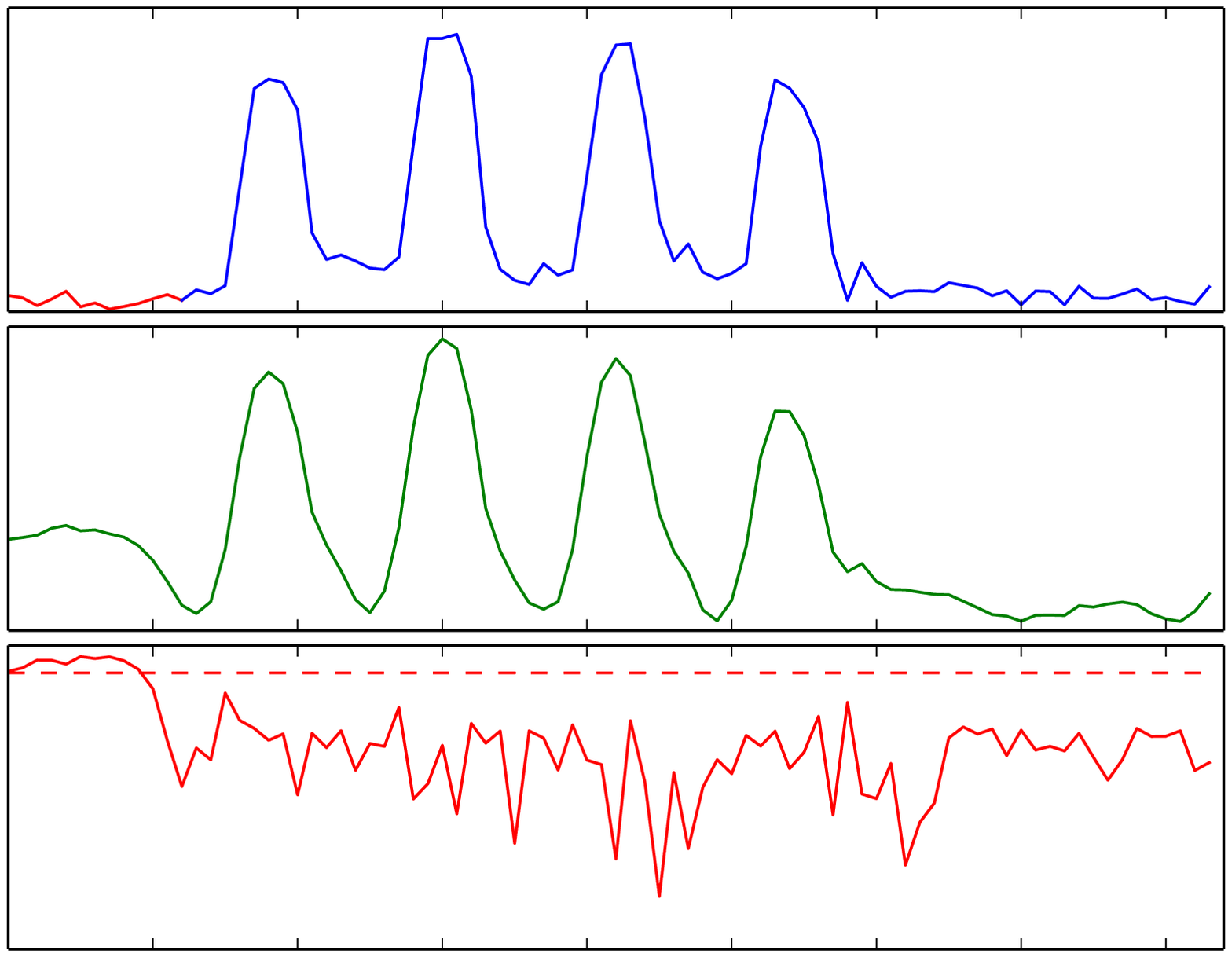}}
    \vspace{-10pt}
    \subfigure[Space Shuttle-N \label{fig:space_n}]{\includegraphics[width=0.495\linewidth]{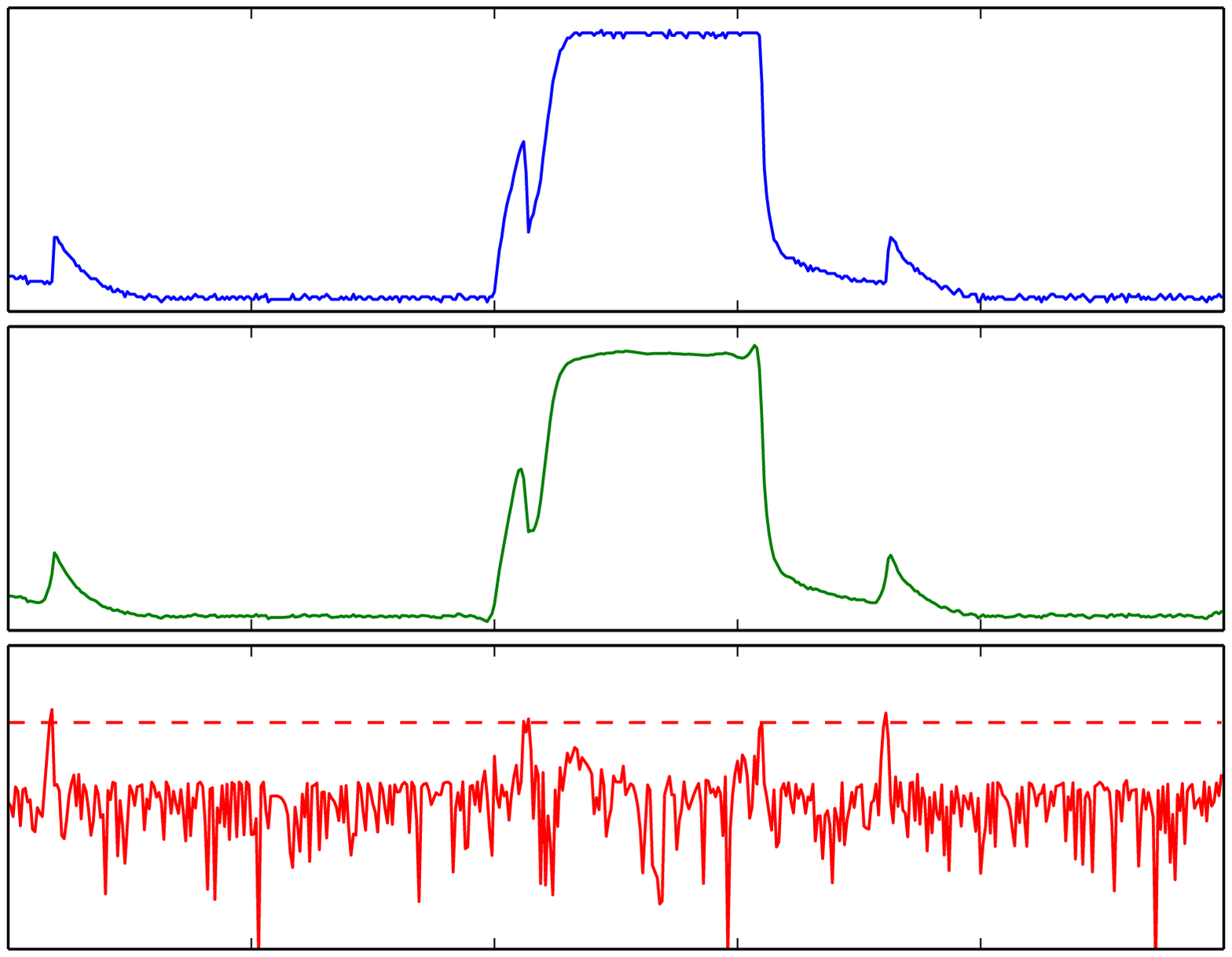}}
    \subfigure[Space Shuttle-A \label{fig:space_a}]{\includegraphics[width=0.495\linewidth]{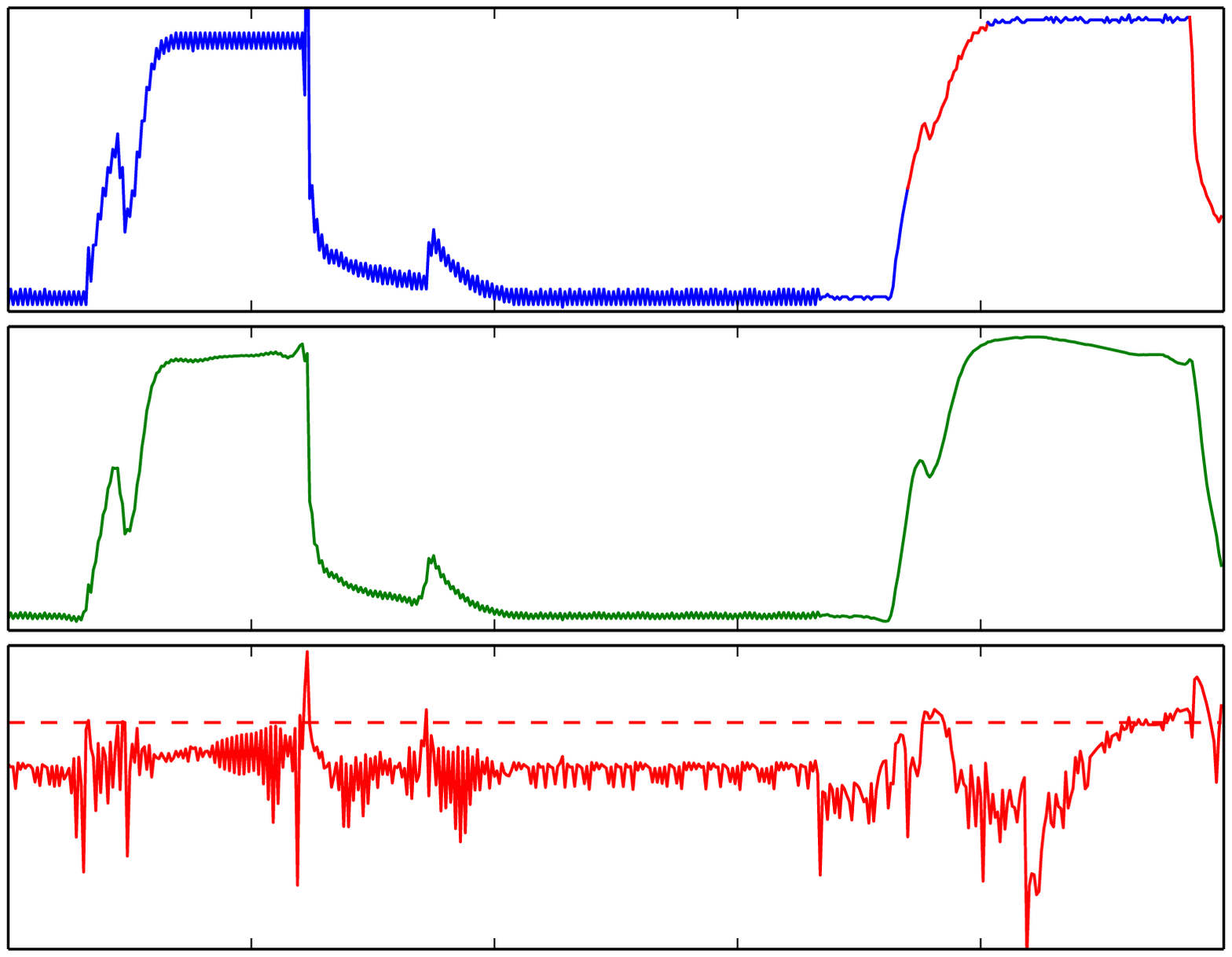}}
    \subfigure[Engine-P-N]{\includegraphics[width=0.495\linewidth]{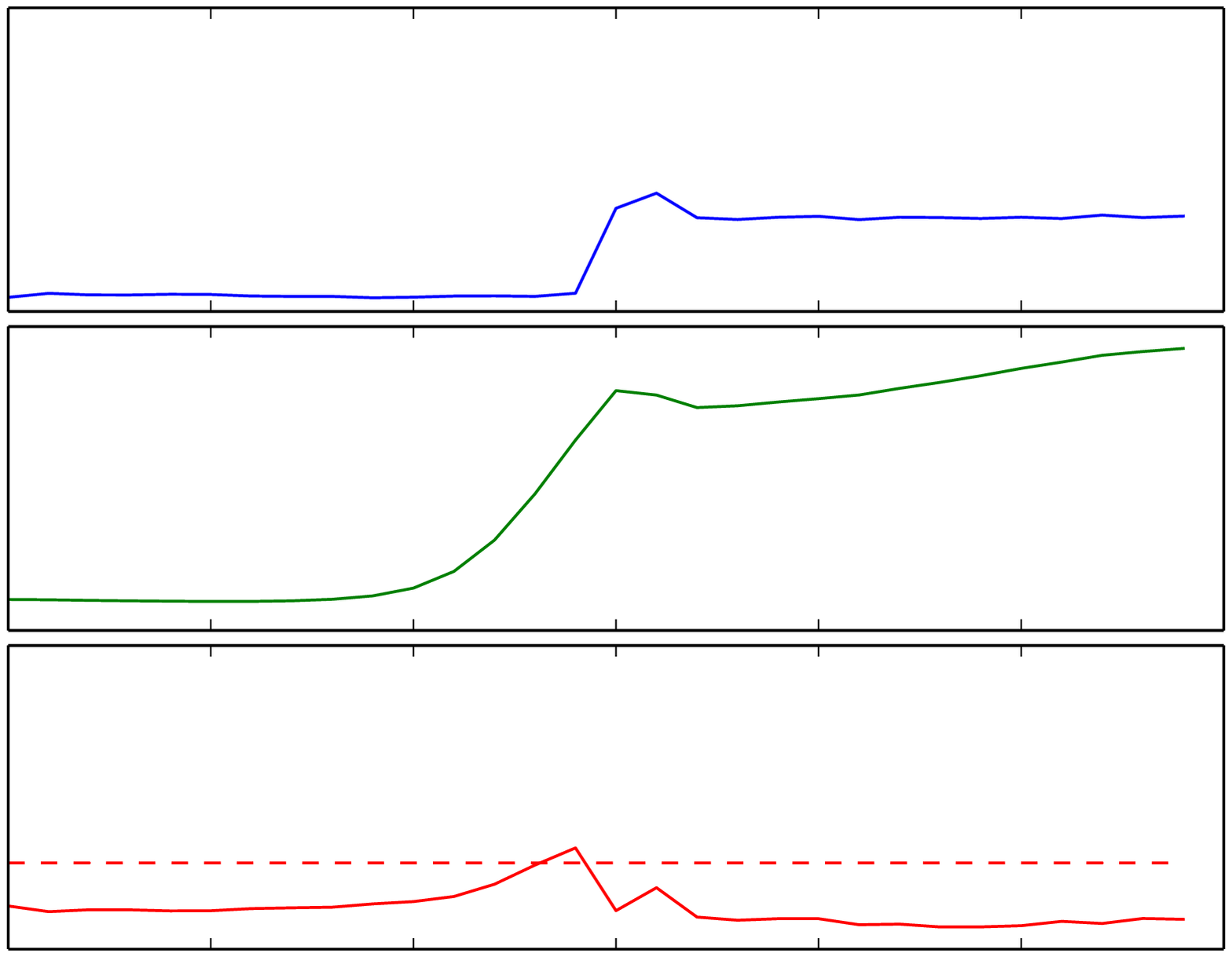}}
    \subfigure[Engine-P-A]{\includegraphics[width=0.495\linewidth]{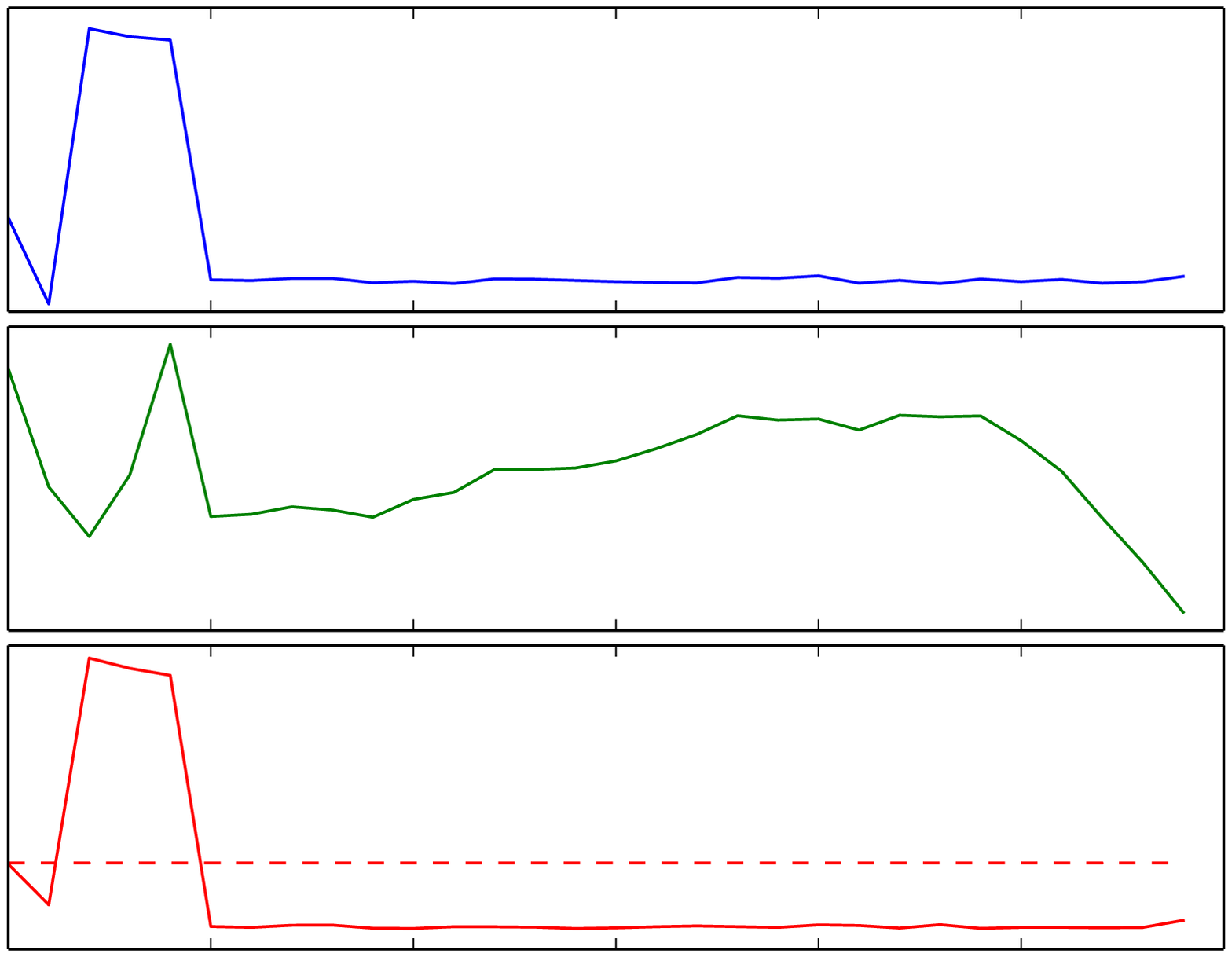}}
    \subfigure[Engine-NP-N]{\includegraphics[width=0.495\linewidth]{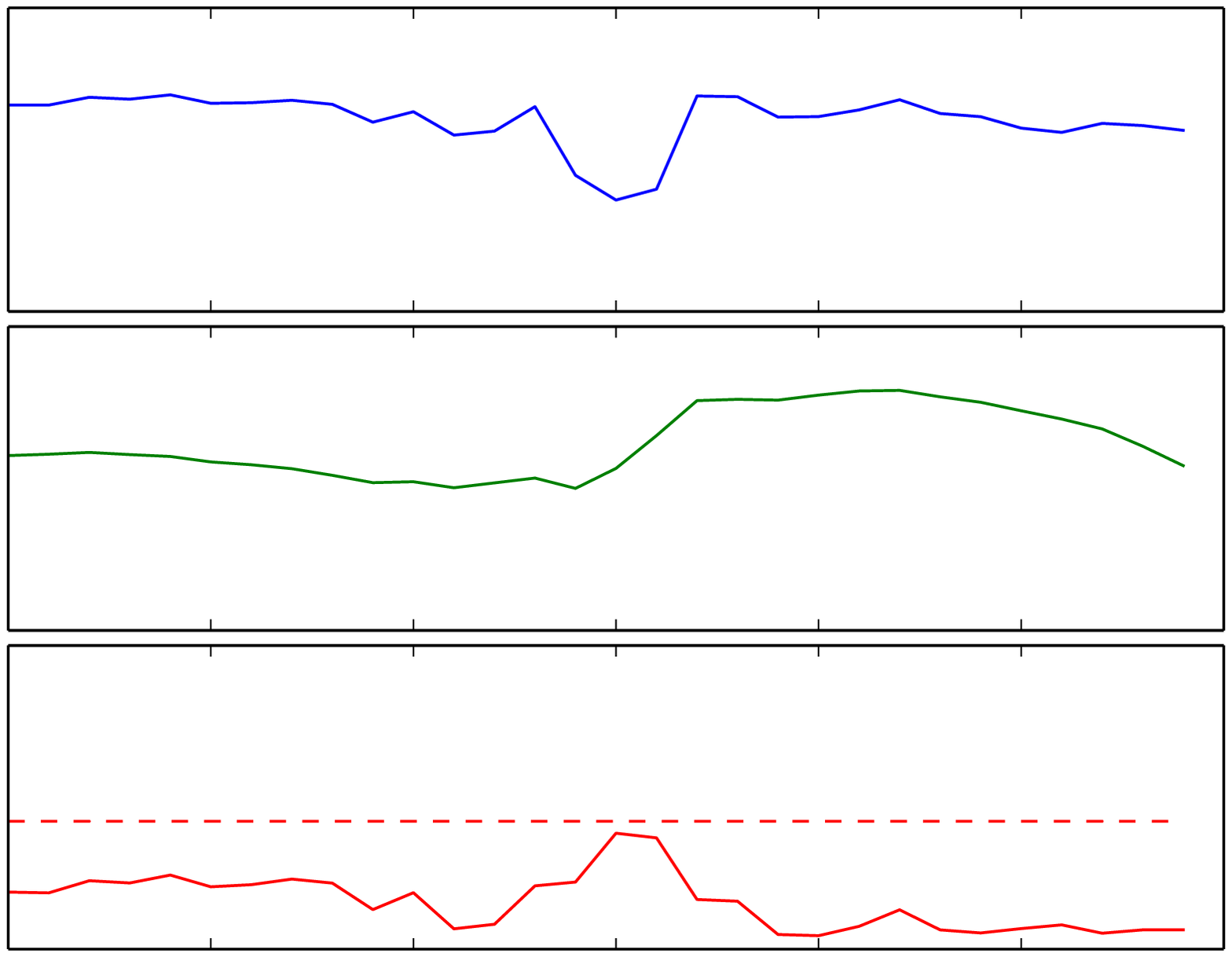}}
    \subfigure[Engine-NP-A]{\includegraphics[width=0.495\linewidth]{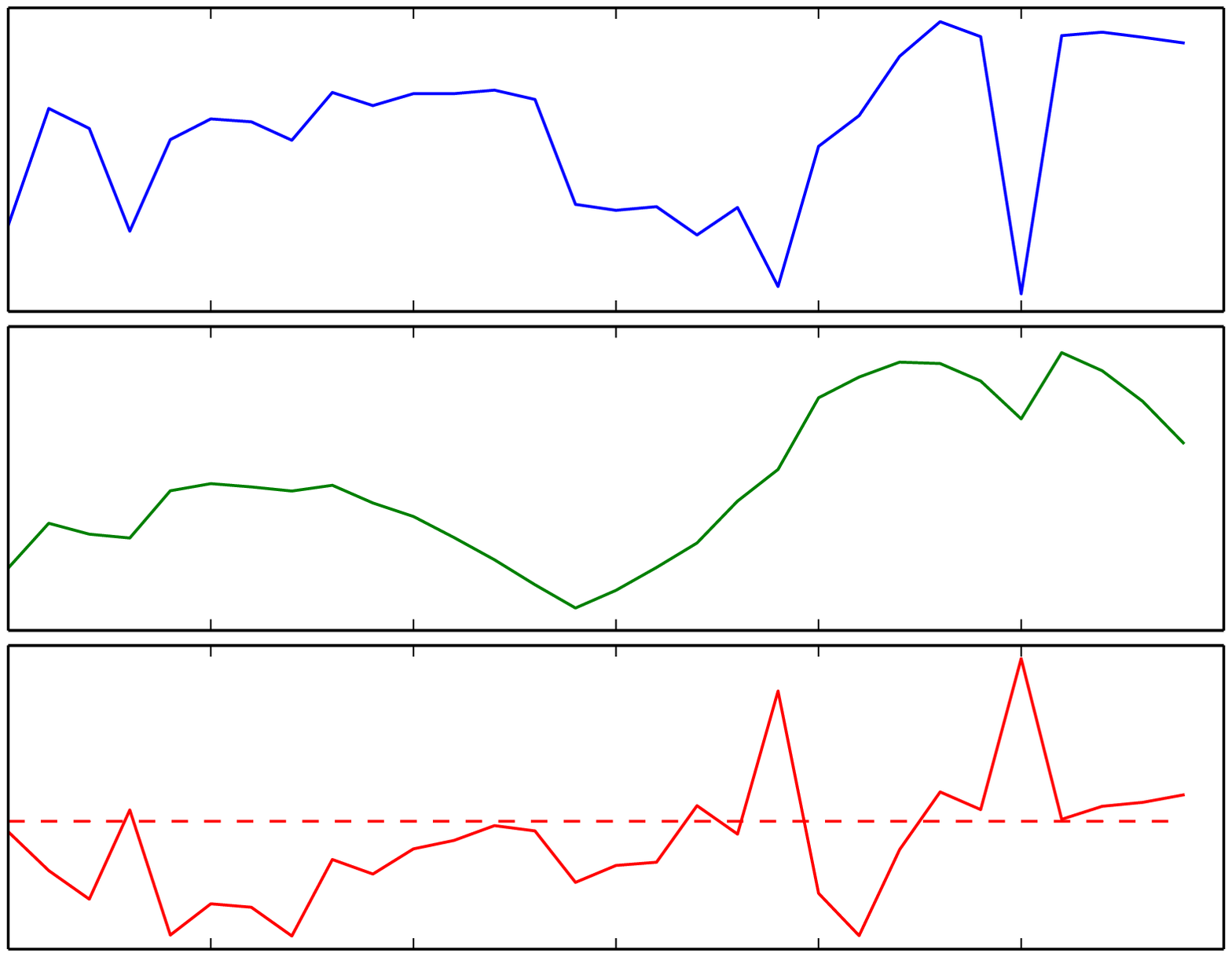}}
    \vspace{-10pt}
    \subfigure[ECG-N]{\includegraphics[width=0.495\linewidth]{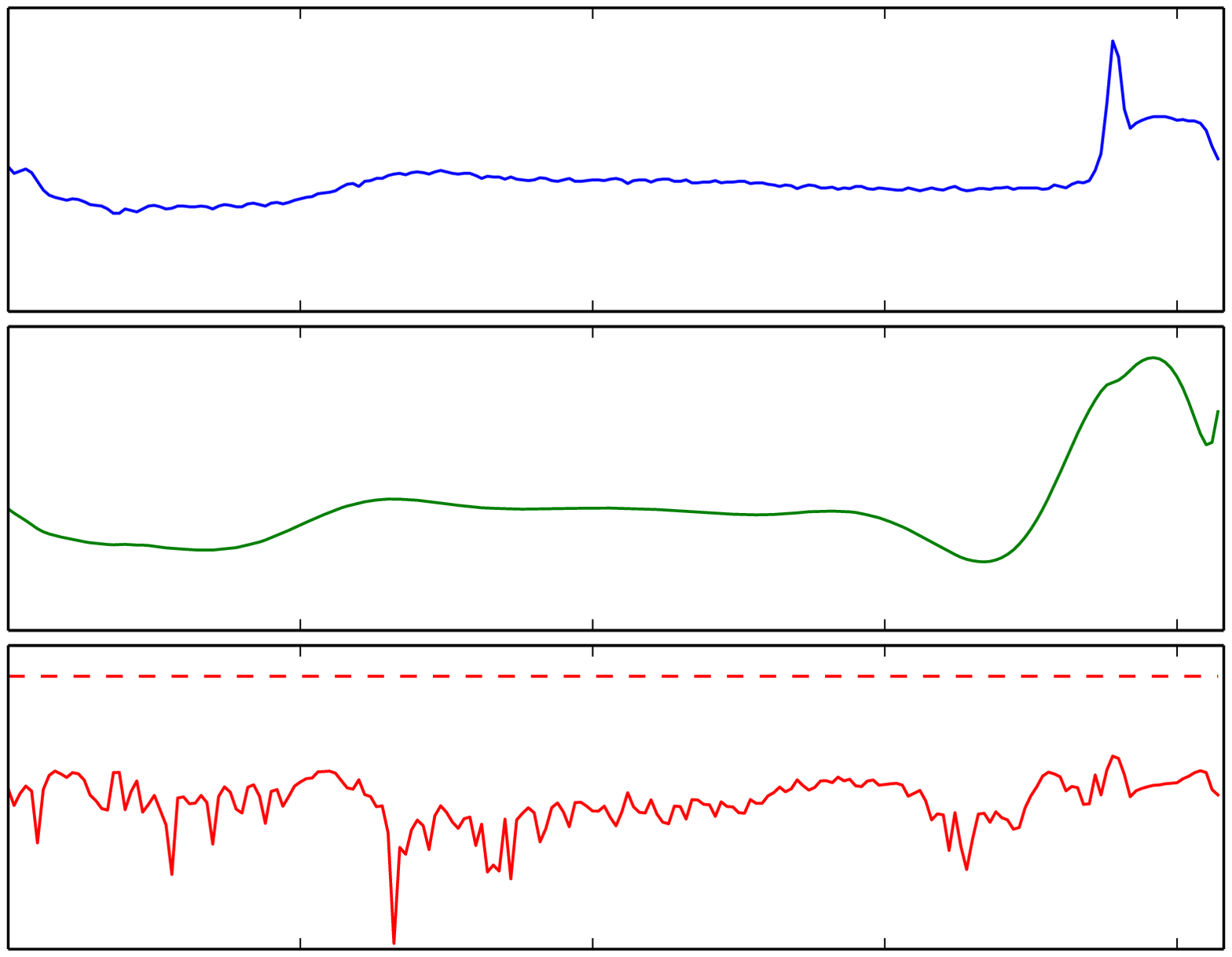}}
    \subfigure[ECG-A\label{fig:ecg_a}]{\includegraphics[width=0.495\linewidth]{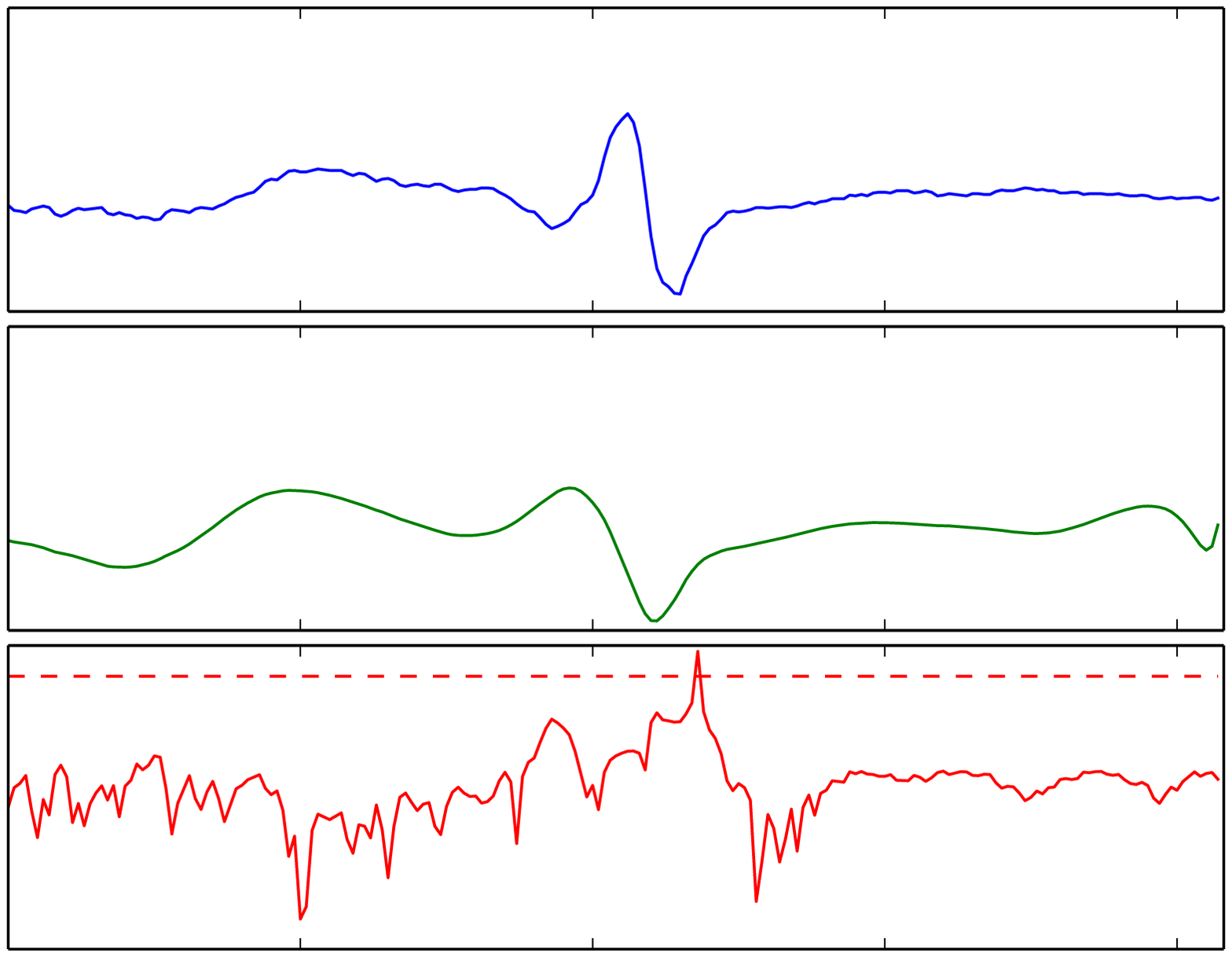}}
    \caption{Sample original normal (first column) and anomalous (second column) sequences (first row, blue color) with corresponding reconstructed sequences (second row, green color) and anomaly scores (third row, red color). The red regions in the original time-series for anomalous sequences correspond to the exact location of the anomaly in the sequence (whenever available). Plots in same row have same y-axis scale. The anomaly scores are on log-scale.}
\end{figure}

\subsection{Datasets}
\textbf{Power demand dataset}
contains one univariate time-series with $35,040$ readings for power demand recorded over a period of one year. The demand is normally high during the weekdays and low over the weekend. Within a day, the demand is high during working hours and low otherwise (see Figure \ref{fig:power_n}, top-most subplot). A week when any of the first $5$ days has low power demands (similar to the demand over the weekend) is considered anomalous (see Figure \ref{fig:power_a} where first day has low power demand).
We downsample the original time-series by $8$ to obtain non-overlapping sequences with $L=84$ such that each window corresponds to one week. 
\newline
\textbf{Space shuttle dataset}
contains periodic sequences with $1000$ points per cycle, and $15$ such cycles. We delibrately choose $L=1500$ such that a subsequence covers more than one cycle (1.5 cycles per subsequence) and consider sliding windows with step size of $500$. We downsample the original time-series by $3$. The normal and anomalous sequences in Figure \ref{fig:space_n}-\ref{fig:space_a} belong to TEK17 and TEK14 time-series, respectively. 
\newline
\textbf{Engine dataset}
contains readings for $12$ sensors such as coolant temperature, torque, accelerator (control variable), etc. We consider two differents applications of the engine: Engine-P and Engine-NP. Engine-P has a discrete external control with two states: `high' and `low'. The resulting time-series are predictable except at the time-instances when the control variable changes. On the other hand, the external control for Engine-NP can assume any value within a certain range and changes very frequently, and hence the resulting time-series are unpredictable. Sample sequences for the control variables from Engine-P and Engine-NP are shown in Figure \ref{fig:predictable} and \ref{fig:unpredictable}, respectively.
We randomly choose $L=30$ for both Engine-P and Engine-NP.
We reduce the multivariate time-series to univariate by considering only the first principal component after applying principal component analysis \cite{book:pca}. The first component captures $72\%$ of the variance for Engine-P and $61\%$ for Engine-NP.
\newline
\textbf{ECG dataset}
contains quasi-periodic time-series (duration of a cycle varies from one instance to another).
For our experiment, we use the first channel from qtdb/sel102 dataset where the time-series contains one anomaly corresponding to a pre-ventricular contraction (see Figure \ref{fig:ecg_a}). We consider non-overlapping subsequences with $L=208$ (each subsequence corresponds to approximately $800$ms). Since only one anomaly is present in the dataset, sets $v_{N2}$ and $v_{A}$ are not created. The best model, i.e. $c$, is chosen based on the minimum reconstruction error on set $v_{N1}$. We choose $\tau=\mu_a+\sigma_a$, where $\mu_a$ and $\sigma_a$ are the mean and standard deviation of the anomaly scores of the points from $v_{N1}$.

\subsection{Observations}
The key observations from our experiments are as follows:
\newline
1) 
The positive likelihood ratio is significantly higher than 1.0 for all the datasets (see Table \ref{t:results}). High positive likelihood ratio values suggest that EncDec-AD gives significantly higher anomaly scores for anomalous points as compared to normal points.
 \newline
2) For periodic time-series, we experiment with varying window lengths: window length same as the length of one cycle (power demand dataset) and window length greater than the length of one cycle (space shuttle dataset). We also consider a quasi-periodic time-series (ECG). EncDec-AD is able to detect anomalies in all these scenarios.
\newline
3) A time-series prediction based anomaly detection model LSTM-AD \cite{p:lstm-ad} gives better results for the predictable datasets: Space Shuttle, Power and Engine-P (corresponding to Engine dataset in \cite{p:lstm-ad}) with $F_{0.1}$ scores of $0.84$, $0.90$ and $0.89$, respectively. On the other hand, EncDec-AD gives better results for Engine-NP where the sequences are not predictable. The best LSTM-AD model gives P, R, $F_{0.05}$ and TPR/FPR of $0.03, 0.07, 0.03, 1.9$, respectively (for a two hidden layer architecture with 30 LSTM units in each layer and prediction length of $1$) owing to the fact that the time-series is not predictable and hence a good prediction model could not be learnt, whereas EncDec-AD gives P, R, $F_{0.1}$ score and TPR/FPR of $0.96, 0.18, 0.93$ and $7.6$, respectively.

\section{Related Work}
Time-series prediction models have been shown to be effective for anomaly detection by using the prediction error or a function of prediction error as a measure of the severity of anomaly \cite{pred2_AD, ma2003online,p:markovAD}. Recently, deep LSTMs have been used as prediction models in LSTM-AD \cite{p:lstm-ad,p:lstm-ad-ecg,p:lstm-ad-ode} where a prediction model learnt over the normal time-series using LSTM networks is used to predict future points, and likelihood of prediction error is used as a measure of anomaly. EncDec-AD learns a representation from the entire sequence which is then used to reconstruct the sequence, and is therefore different from prediction based anomaly detection models. Non-temporal reconstruction models such as denoising autoencoders for anomaly detection \cite{p:autoencoderAD} and Deep Belief Nets \cite{p:dbn-eeg-AD} have been proposed. For time-series data, LSTM based encoder-decoder is a natural extension to such models. 

\section{Discussion}
We show that LSTM Encoder-Decoder based reconstruction model learnt over normal time-series can be a viable approach to detect anomalies in time-series. Our approach works well for detecting anomalies from predictable as well as unpredictable time-series. Whereas many existing models for anomaly detection rely on the fact that the time-series should be predictable, EncDec-AD is shown to detect anomalies even from unpredictable time-series, and hence may be more robust compared to such models. 
The fact that EncDec-AD is able to detect anomalies from time-series with length as large as $500$ suggests the LSTM encoder-decoders are  learning a robust model of normal behavior.

\bibliography{esann,nips2015,icml2016}
\bibliographystyle{icml2016}

\end{document}